# A New 27 Class Sign Language Dataset Collected from 173 Individuals

Arda Mavi     Zeynep Dikle

*Ayrancı Anadolu High School, Ankara, Turkey*

**Abstract**

After the interviews, it has been comprehended that speech-impaired individuals who use sign languages have difficulty communicating with other people who do not know sign language. Due to the communication problems, the sense of independence of speech-impaired individuals could be damaged and lead them to socialize less with society. To contribute to the development of technologies, that can reduce the communication problems of speech-impaired persons, a new dataset was presented with this paper. The dataset was created by processing American Sign Language-based photographs collected from 173 volunteers, published as "27 Class Sign Language Dataset" on the Kaggle Datasets web page.

Dataset is available at https://www.kaggle.com/ardamavi/27-class-sign-language-dataset upon publication of this paper.

## 1 Introduction

One of the communication methods of speech-impaired people is sign language, moreover, the American Sign Language (ASL) is being used by approximately 250,000 to 500,000 Americans and also some Canadians [1]. Communication problems between speech-impaired individuals and the people who do not know sign language have been one of the main focuses in the interviews made with speech-impaired individuals and people who work with speech-impaired individuals. Communication difficulties may cause speech-impaired people to alienation from society, and damage their sense of independence due to the inability of communicating alone with other people, who do not know sign language. Although various studies have been carried out to solve the communication problems, like online sign language courses [2] and sign language dictionaries [3], especially computer vision studies [4,5,6] have been a prior motivation for this work. Image classification algorithms, Convolutional Neural Networks (CNN) [4] have the ability to classify various objects in photographs [5]. The datasets have been used for the preparation (training) of the algorithms; especially the datasets that have the large number of data samples with diversity can increase the classification accuracy of the algorithms [4,5,6]. The promising CNN studies have been an inspiration for collecting a new sign language dataset. With this work, it was aimed to prepare a new sign language dataset that can meet the diversity and sample size necessity of intelligent computer vision studies and the applications of sign language applications.

## 2 Preparation of the Dataset

### 2.1 Data Collection

The data collection and processing plans were submitted to the Institutional Review Board (IRB) in order to obtain the necessary permissions for data collection. After the approval of the IRB, the data collection procedure was started with the consent of 173 volunteers.

26 classes contain numbers, letters, and words selected from American Sign Language, which has the large number of users [1], to gathering the data.

Selected numbers are 0, 1, 2, 3, 4, 5, 6, 7, 8, and 9.

Selected letters are A, B, C, D, and E.

Selected expressions are Hello, Yes, No, Good, Bye, Good morning, Whats up, Pardon, Project, Little bit, and Please.

Images of these selected classes were shown to 173 volunteers and they were asked to do the sign language gesture, with their right hand. The gestures were taken with a camera with 3024 pixels to 3024 pixels frame size and RGB color space. A total of 130 photographs were taken from each individual, 5 from each class (minor changes on sample sizes in classes can be observed). All the shots have different angles with minor changes. Sign language gestures were placed approximately in the middle of the frame. Photos were taken in different places to provide diversity for background and light. Artificial environment and light equipment were not used during the data collection. Sample photos from the collected data are shown in Figure 1.

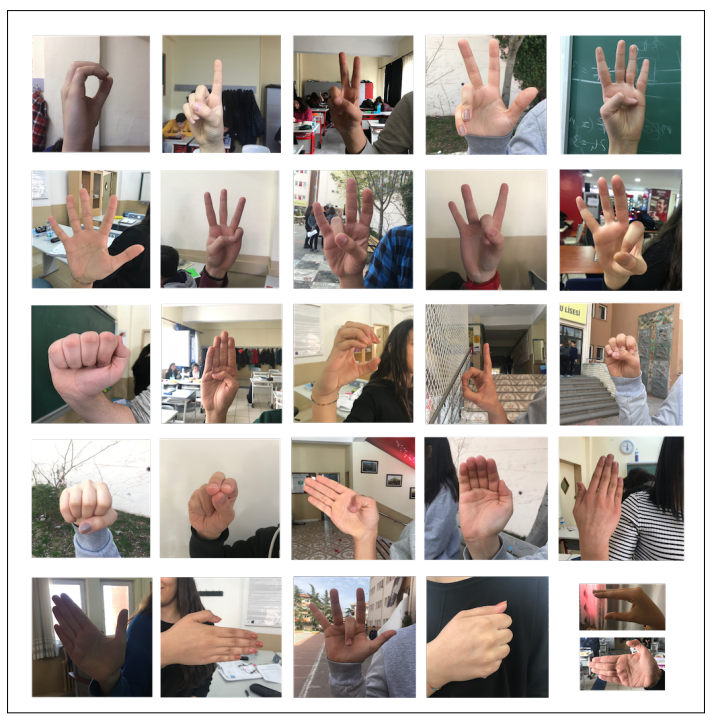

**Figure 1.** Sample Photos From Collected Data

**2.2 NULL Class**

The NULL class contains the randomly taken 314 photos in the environments where the data was collected. Photographs belonging to this class were collected without any sign language gestures in the frame. With the purpose of enabling the dataset users to control and/or improve their projects, the NULL class was included as the 27th class in the dataset.

**2.3 Data Processing**

Version 3.6.0 of the Python [7] programming language and version 3.2.0-dev of the OpenCV [8] image processing library was used to process the data. The collected photos were resized to 128 pixels to 128 pixels frame size to minimize the size of the data without affecting the intelligibility of the gesture in the frame, and avoid personal data sharing such as hand and fingerprint. Integers pixel values in the range of [0, 255] were normalized to the 32-bit floats in the range of [0, 1].

**2.4 Data Storage**

Version 1.19.2 of the NumPy [9] library was used to combine and store the data. The processed photos were combined into a 4-dimensional NumPy tensor named "X" with a form of (n, 128, 128, 3), which represents the number of photos (n), photo height and width (128x128 px), and three color channels (RGB). Letters, characters, or expressions corresponding to the sign language gesture in each photograph converted into a String (U17) data type with the lowercase characters and combined into a 2-dimensional NumPy tensor named "Y" in the form of (n, 1), which n represents the same number of photos like tensor X. For example, the sign language gesture in the fifth (n=5) element of the tensor X matched with the fifth element (n=5) of the tensor Y. The label "NULL" represents the Null class in the tensor Y. The created NumPy tensors X and Y were saved with the file extension ".npy", which represents the NumPy file. The tensor X, which has shape (22801, 128, 128, 3), has the file name "X.npy" and the tensor Y, which has shape (22801, 1), has the file name "Y.npy".

Classes (labels) in the tensor Y are '0', '1', '2', '3', '4', '5', '6', '7', '8', '9', 'a', 'b', 'c', 'd', 'e', 'bye', 'good', 'good morning', 'hello', 'little bit', 'no', 'pardon', 'please', 'project', 'whats up', 'yes', and 'NULL'.

**3 Publishing the Dataset**

Kaggle [10] platform was chosen to share the processed dataset. After the project, named "27 Class Sign Language Dataset", was created on the Kaggle Dataset platform, prepared "X.npy" and "Y.npy" files were uploaded to the project. The dataset project will be publicly available at https://www.kaggle.com/ardamavi/27-class-sign-language-dataset upon publication of the paper. The dataset can be used for research and/or commercial purposes by citing the paper.

# 5 Conclusion and Future Works

The problems, experienced by speech-impaired individuals using sign language in communicating with people who do not know sign language, were examined. Investigations have been made for possible technological solutions for the communication problem. A dataset preparation plan was created to contribute to the development of intelligent computer vision algorithms [4,5,6] and their application for sign language. In order to create the diversity of data needed in the preparation of computer vision algorithms [4,5,6], photos of 26 different sign language gestures were collected from 173 volunteers in front of different backgrounds, light directions, and minor angle changes. Considering the control and development purpose, 314 random photos, that do not contain sign language gestures, were added to the dataset and labeled as "NULL" class. Collected data were processed to facilitate the usage of the dataset. The dataset project web page will be publicly available at https://www.kaggle.com/ardamavi/27-class-sign-language-dataset upon publication of the paper. In future studies, new data can be collected and added to the dataset. Considering the importance of the angle of sign language gesture position, data augmentation [11] methods can also be applied in order to increase the number of data. With this study, new technologies can be developed that can facilitate the communication of people who use sign language and who do not know sign language.

**Acknowledgements**

Thanks to all of the volunteer students and teachers from Ayrancı Anadolu High School for their help in collecting data, and Zümra Uğur, Demet Coşkun, and Kadir Efe Dilek who contributions to the research on speech-impaired people.

## *References*

[1] Mitchell, R. E., Young, T. A., Bachleda, B., and Karchmer, M. A. (2006). How Many People Use ASL in the United States? Why Estimates Need Updating. *Sign Language Studies*, volume 6, number 3.

[2] Communication Services for the Deaf and Hard of Hearing. (2022). American Sign Language Classes. *csdhh.org* (Accessed: March 7, 2022)

[3] Sternberg, M. L. A. (1981). American Sign Language Dictionary. *Harper Perennial*.

[4] LeCun, Y., Bottou, L., Bengio, Y., and Haffner, P. (1998). Gradient-based learning applied to document recognition. *Proceedings of the IEEE*, volume 86, number 11, pages 2278-2324.

[5] Krizhevsky, A., Sutskever, I., and Hinton, G. E. (2017). ImageNet Classification with Deep Convolutional Neural Networks. *Advances in neural information processing systems*, 25.

[6] Simonyan, K., and Zisserman, A. (2014). Very deep convolutional networks for large-scale image recognition. *arXiv*. Preprint.

[7] Van Rossum, G. (1995). Python Tutorial. *Centrum voor Wiskunde en Informatica*.

[8] Bradski, G. (2000). The OpenCV Library. *Dr. Dobb's Journal of Software Tools*.

[9] Harris, C.R., Millman, K.J., van der Walt, S.J. et al. (2020). Array programming with NumPy. *Nature* 585, 357–362.

[10] Kaggle Inc. (2010). Kaggle. *kaggle.com* (Accessed: March 7, 2022)

[11] Mikołajczyk, A., and Grochowski, M. (2018). Data augmentation for improving deep learning in image classification problem. *2018 international interdisciplinary PhD workshop*. IEEE, pages. 117-122.